\documentclass[sigconf]{acmart}
\AtBeginDocument{%
  }

\copyrightyear{2026}
\acmYear{2026}
\setcopyright{cc}
\setcctype{by}
\acmDOI{10.1145/3767308.3834931}
\acmConference[MM '26] {Proceedings of the 34th ACM International Conference on Multimedia}{November 10--14, 2026}{Rio de Janeiro, Brazil}
\acmBooktitle{Proceedings of the 34th ACM International Conference on Multimedia (MM '26), November 10--14, 2026, Rio de Janeiro, Brazil}
\acmISBN{979-8-4007-2213-4/2026/11}

\acmSubmissionID{mfp0198}

\usepackage{amsthm}
\usepackage{mdframed}
\usepackage{multirow}   
\usepackage[table]{xcolor}  
\usepackage{graphicx}     
\usepackage{subcaption}   
\usepackage{float}
\usepackage{placeins}
\usepackage{diagbox}
\usepackage{comment}
\usepackage{caption}

\newcommand{\sysname}{BioPro}
\newtheorem{lemma}{Lemma}
\begin{document}

\title{\sysname{}: Towards Difference-Aware Gender Fairness for Vision-Language Models}

\author{Yujie Lin}
\authornote{Both authors contributed equally to this research.}
\affiliation{%
  \department{School of Informatics}
  \institution{Xiamen University}
  \city{Xiamen}
  \country{China}}
\email{linyujie@stu.xmu.edu.cn}

\author{Jiayao Ma}
\authornotemark[1]
\affiliation{%
  \department{School of Informatics}
  \institution{Xiamen University}
  \city{Xiamen}
  \country{China}}
\email{majiayao@stu.xmu.edu.cn}

\author{Qingguo Hu}
\affiliation{%
  \department{School of Informatics}
  \institution{Xiamen University}
  \city{Xiamen}
  \country{China}}
\email{huqingguo@stu.xmu.edu.cn}

\author{Wenbo Li}
\affiliation{%
  \department{School of Informatics}
  \institution{Xiamen University}
  \city{Xiamen}
  \country{China}}
\email{liwenbo2099@outlook.com}

\author{Genji Li}
\affiliation{%
  \institution{Taowhale Lab}
  \city{Xiamen}
  \country{China}}
\email{genji@taowhale.com}

\author{Derek Wong}
\affiliation{%
  \department{Department of Computer and Information Science}
  \institution{University of Macau}
  \city{Macau}
  \country{China}}
\email{derekfw@um.edu.mo}

\author{Jinsong Su}
\authornote{Corresponding author.}
\affiliation{%
  \institution{Key Lab of Digital Protection and Intelligent Processing of Intangible Cultural Heritage of Fujian-Taiwan (XMU), Ministry of Culture and Tourism}
  \city{Xiamen}
  \country{China}}
\email{jssu@xmu.edu.cn}

\renewcommand{\shortauthors}{Yujie Lin et al.}

\begin{abstract}
Vision-Language Models (VLMs) inherit significant social biases from their training data, notably in gender representation. Current fairness interventions often adopt a ``difference-unaware'' perspective that enforces uniform treatment across demographic groups. These approaches, however, fail to distinguish between contexts where neutrality is required and those where group-specific attributes are legitimate and must be preserved. Building upon recent advances in difference-aware fairness for text-only models, we extend this concept to the multimodal domain and formalize the problem of difference-aware gender fairness for image captioning and text-to-image generation. We advocate for selective debiasing, which aims to mitigate unwanted bias in neutral contexts while preserving valid distinctions in explicit ones. To achieve this, we propose \sysname{} (Bias Orthogonal Projection), an entirely training-free framework. \sysname{} identifies a low-dimensional gender-variation subspace through counterfactual embeddings and applies projection to selectively neutralize gender-related information. Experiments show that \sysname{} effectively reduces gender bias in neutral cases while maintaining gender faithfulness in explicit ones, thus providing a promising direction toward achieving selective fairness in VLMs. Beyond gender bias, we further demonstrate that BioPro can effectively generalize to continuous bias variables, such as scene brightness, highlighting its broader applicability. 
\end{abstract}



\maketitle

\section{Introduction}
\label{sec:intro}
\begin{figure}[!t]
    \centering
    \includegraphics[width=\linewidth]{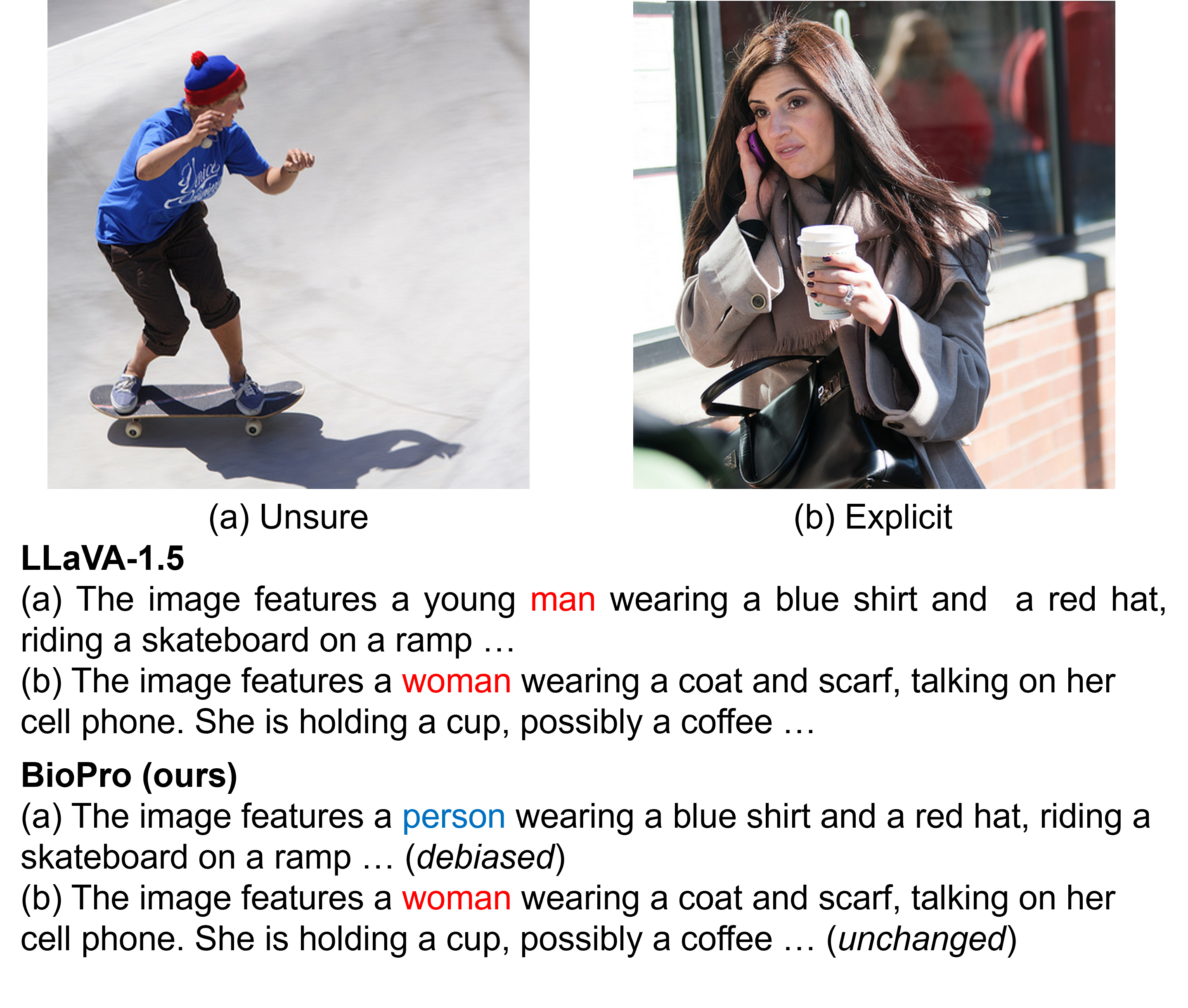} 
    \vspace{-8mm}
    \caption{\sysname{} can selectively perform debiasing primarily on neutral samples labeled as unsure gender.
}
    \Description{Comparison of image-captioning outputs before and after selective debiasing for gender-neutral and gender-explicit images.}
    \label{fig: captioning}
    \vspace{-5mm}
\end{figure}
\begin{figure*}[!t]
    \centering
    \includegraphics[width=0.85\linewidth]{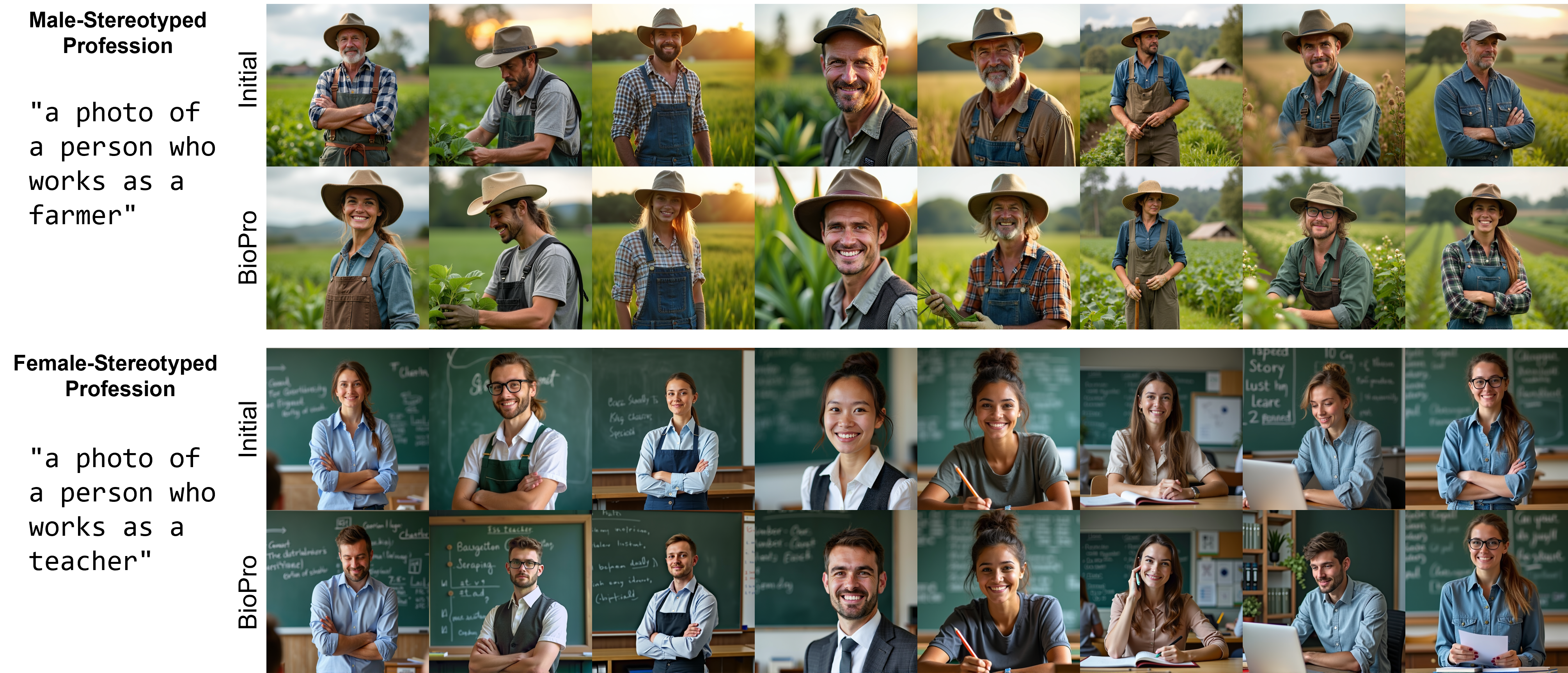}
        \caption{An example of text-to-image generation. The figure shows eight images generated from the same prompt with fixed seeds ranging from 0 to 7. \sysname{} can help balance the gender distribution of the generated images.
}
    \Description{Eight generated portraits for the same occupation prompt, illustrating a more balanced gender distribution after applying BioPro.}
    \label{fig: generation}
    \vspace{-4mm}
\end{figure*}

Large Language Models (LLMs)~\cite{achiam2023gpt,dubey2024llama} have demonstrated remarkable capabilities across diverse natural language tasks. However, their impressive fluency often comes with a significant drawback: \text{social bias}~\cite{nadeem2020stereoset,lin2024towards,lin2025fade,lin2026bidirectional}. These models inherit and amplify patterns of bias from their large-scale training data, leading to unfair or stereotypical outputs~\cite{liang2021towards,parrish2021bbq,gallegos2024bias,shao2024supervised}. Existing fairness research on LLMs has largely adopted a \textit{difference-unaware} perspective, where fairness is operationalized as treating all demographic groups identically regardless of context. While such approaches can reduce overt disparities, they often neglect the nuanced circumstances where group differences are legitimate or even necessary to recognize~\cite{wang2025fairness}.

\text{Difference-aware fairness} offers an alternative perspective that explicitly acknowledges the importance of contextual differentiation. As argued in~\cite{wang2025fairness}, fair behavior does not always entail uniform treatment across groups: in many real-world situations, fair outcomes require sensitivity to social, cultural, or legal distinctions. To this end, they introduce the notion of \textit{difference awareness}, emphasizing that models should discern when it is contextually appropriate to treat groups differently (e.g., gender-specific legal exemptions) and when uniformity is required. Through a comprehensive benchmark suite, they demonstrate that current LLMs  though highly capable and well-aligned, often exhibit \text{difference unawareness}, failing to recognize meaningful group distinctions when they matter most.

While prior studies have focused on text-only settings, the same challenge persists and arguably intensifies in vision language models (VLMs) \cite{liu2023visual,zhu2023minigpt,rombach2022high}. VLMs jointly process multimodal information to perform tasks such as image captioning and text-to-image generation. In multimodal scenarios, fairness entails not only generating unbiased language or images but doing so in a way that is sensitive to visual or textual context. As shown in Fig.~\ref{fig: captioning}, when describing a person in a photo, a captioning model should avoid inferring gender for gender-ambiguous subjects, yet should faithfully preserve gender information when it is visually evident. 
Similarly, the problem can be extended to text-to-image generation. A neutral prompt (e.g., ``\textit{a photo of a person who works as a doctor}'') should yield a balanced gender distribution (Fig.~\ref{fig: generation}), whereas an explicit prompt (e.g., ``\textit{a photo of a female doctor}'') should respect the specified gender identity.
Motivated by these problems, we propose the notion of \text{difference-aware gender fairness }for VLMs, extending the concept of difference awareness from the text domain to multimodal learning. Our goal is to achieve \text{selective debiasing}, mitigating unwanted gender bias in neutral contexts while preserving intended gender semantics in explicit contexts. To this end, we formalize the problem of difference-aware fairness on two representative VLM tasks: \text{image captioning} and \text{text-to-image generation}. For each task, we define a set of constraints that balance three desirable properties: (i) neutral fairness (bias suppression for neutral inputs), (ii) explicit gender faithfulness (gender consistency for gender-explicit cases), and (iii) semantic preservation (maintaining task-relevant content).

Building on this formulation, we introduce a novel debiasing framework termed \textbf{Bi}as \textbf{O}rthogonal \textbf{Pro}jection (\textbf{\sysname{}}), which enables targeted removal of bias-related components from multimodal representations. The key idea is to identify a low-dimensional \text{gender-variation subspace} through paired counterfactual embeddings, and then perform orthogonal projection to selectively neutralize gender-related information. To avoid over-correction, we further introduce a \text{projection-based selection mechanism} that adapts debiasing strength according to the degree of gender expression in each sample. For text-to-image generation, where the output inherently contains gender attributes, we propose an additional \text{calibration term} to balance gender proportions in neutral prompts without erasing intended semantics in explicit ones. 
In particular, without the difference-aware setting, our method can generalize to other types of bias. \sysname{} is capable of handling non-discrete, continuous bias variables. For instance, when generating images of forests, the model tends to produce scenes with sufficient illumination while lacking samples under dim lighting conditions. In this case, environmental brightness serves as a continuous bias variable, which we refer to as \text{scene bias}. \sysname{} can effectively control the model to generate more scenes with lower brightness, and the illumination level can be flexibly adjusted through a hyperparameter.
In summary, our contributions are three-fold:
\begin{itemize}
    \item We firstly extend the concept of \text{difference-aware fairness} from LLMs to VLMs, providing a principled formulation for selective bias mitigation in multimodal tasks.
    \item We propose \sysname{}, an entirely \textbf{training-free} framework for both image captioning and generation, ensuring fairness without compromising semantic fidelity.
     To the best of our knowledge, \sysname{} is the first method that introduces and effectively handles continuous bias variables in text-to-image generation tasks.
    \item Through comprehensive experiments, we demonstrate that \sysname{} effectively reduces gender bias in neutral cases while maintaining gender faithfulness in explicit ones, highlighting the feasibility of selective fairness in vision–language systems. Code is available at \url{https://github.com/XMUDeepLIT/BioPro}.
\end{itemize}


\section{Related Works}

\textbf{Difference-aware Bias in LLMs.}
Recent work \cite{wang2025fairness} challenges the notion of “fairness as blindness.” They formalize difference-aware fairness, arguing that models must distinguish between spurious stereotypes (e.g., nurse $\rightarrow$ female) and legitimate, context-regulated differences (e.g., gender-specific medical descriptions). Moreover, \citet{wang2025fairness} show that common debiasing techniques, such as moral self-correction prompts~\cite{ganguli2023capacity, pan2024automatically,liu2024intrinsic}, tend to {reduce} difference awareness. Their results underscore a fundamental limitation of current bias mitigation paradigms: striving for demographic parity can erase legitimate contextual distinctions and paradoxically produce less fair outcomes in sensitive applications.

\textbf{Social Bias in VLMs.} 
VLMs inevitably inherit social stereotypes from large-scale vision–language corpora, leading to reliance on contextual shortcuts (e.g., kitchen → woman)~\cite{zhao2017men, Hendricks2018Women} and amplification of such biases during generation \cite{hirota2022quantifying}. Although fine-tuning based debiasing strategies \cite{Zhang2023iti, shen2024finetuning, li2025fairmapping, li2024selfdiscovery, kim2023destereotyping, Gandikota2024concept,lin2026zerounlearn,jia2026object} exist, they are often costly and model-specific, motivating growing interest in post-hoc (inference-time) interventions for their flexibility and generality. Such methods generally fall into three categories: (i) prompt-space interventions \cite{berg2022prompt, friedrich2023fairdiffusion, sahili2025faircot, kim2023destereotyping}; (ii) representation-space interventions (e.g., via projection) \cite{jung2024unified, gerych2024bendvlm, chuang2023debiasing, Seth2023DeAR, fu2025fairimagen}; and (iii) post-generation corrections, which adjust generated outputs after inference \cite{hirota2023model, narayanan2025bias}.
As a representative work in the second category of debiasing methods, \cite{bolukbasi2016man} mitigates bias in text-only representations by removing a predefined gender direction. In contrast, \sysname{} operates on fused text–image representations, identifies a bias subspace from the principal bias-relevant directions, and performs debiasing through orthogonal projection onto its complement.
\section{Difference-Aware Fairness for VLMs}
In this section, we define the difference-aware gender fairness problem on two representative vision-language tasks: image captioning and text-to-image generation. 
The subscripts $c$ and $g$ denote captioning and generation, respectively. 
\label{sec: background}
\subsection{Image Captioning}
Consider an input image $x\in \mathcal{X}$ and a text prompt $t_c$. 
Given a VLM $M_c$, we obtain the caption of $x$ as $y_c = M_c(x, t_c)$. 
For the input image set $\mathcal{X}$, we divide it into two subsets:
the {explicit set} $\mathcal{X}_e$, which contains images with clearly identifiable gendered subjects,
and the {neutral set} $\mathcal{X}_n = \mathcal{X} ~\backslash~ \mathcal{X}_e$, 
which includes images whose subjects’ gender is ambiguous or difficult to determine. 
For each image $x$, we extract a multimodal joint embedding:  $h(x, t_c)$.
We then introduce a debiasing transformation $F_c$ acting on $h(x, t_c)$:

    $\tilde{h}(x, t_c) = F_c\big(h(x, t_c)\big)$,
where $\tilde{h}$ represents the debiased multimodal representation.
The desired properties of $F_c$ are as follows:

\textbf{(i) Neutral Fairness.} For a neutral image $x_n \in \mathcal{X}_n$, 
the caption generated from the debiased embedding, $M_c(\tilde{h}(x_n, t_c))$, 
should be free from gender-indicative words $w_g$ (e.g., ``\textit{man}'', ``\textit{woman}'', ``\textit{male}'', ``\textit{female}''):
\begin{equation}
    \mathbb P\big(w_g \mid M_c(\tilde{h}(x_n, t_c))\big) \approx 0.
    \label{eq:gender_constraint}
\end{equation}

\textbf{(ii) Explicit Gender Faithfulness.} 
For an explicit image $x_e \in \mathcal{X}_e$ whose subject gender is clearly identifiable (e.g., male or female), 
the debiasing transformation should preserve the original gender semantics during caption generation.
Formally, the probability of generating the gender word $w_g$ consistent with the image should remain unchanged:
\begin{equation}
    \frac{\mathbb P(w_g \mid M_c(\tilde{h}(x_e, t_c)))}{\mathbb P(w_g \mid M_c(h(x_e, t_c)))}
    \approx 1
    .
    \label{eq:explicit_gender_consistency}
\end{equation}

\textbf{(iii) Semantic Preservation.} For an arbitrary image $x \in \mathcal{X}$, 
the debiasing transformation should not significantly alter the semantics of the original representation:
\begin{equation}
    d\big(\tilde{h}(x, t_c), h(x, t_c)\big) \le \epsilon_c,
    \label{eq:semantic_constraint}
\end{equation}
where $d(\cdot,\cdot)$ denotes a distance metric and $\epsilon_c$ is a small tolerance constant. Therefore, the goal of difference-aware fairness in image captioning is to learn a transformation
$F_c: h(x,t_c) \mapsto \tilde{h}(x,t_c)$ that simultaneously satisfies
Neutral Fairness (Eq.~\ref{eq:gender_constraint}),
Explicit Gender Faithfulness (Eq.~\ref{eq:explicit_gender_consistency}),
and Semantic Preservation (Eq.~\ref{eq:semantic_constraint})
over the corresponding subsets $\mathcal{X}_n$, $\mathcal{X}_e$, and $\mathcal{X}$.
\begin{figure*}[t]
  \centering
  \begin{subfigure}[t]{0.45\textwidth}
    \centering
    \includegraphics[width=\textwidth]{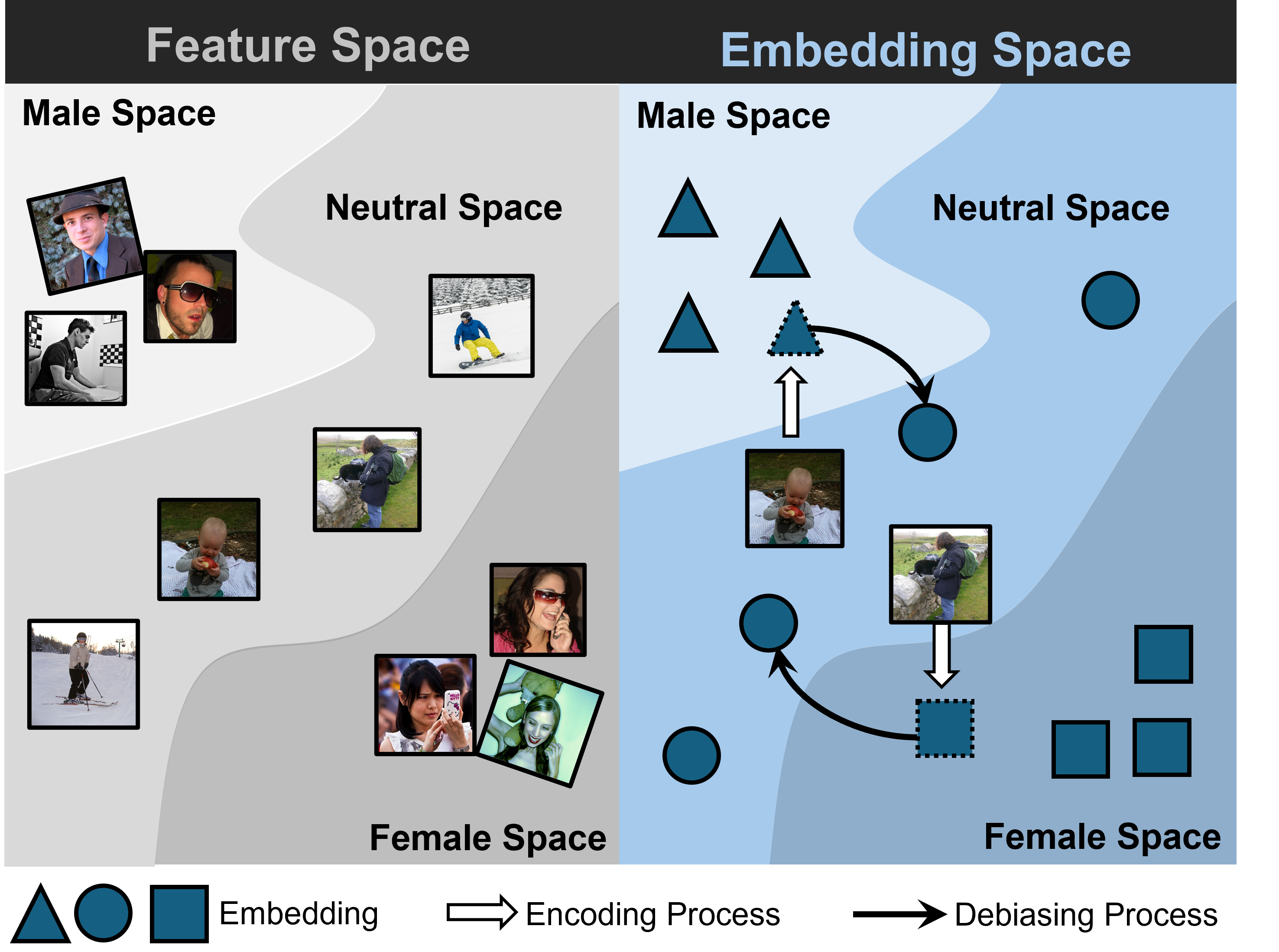}
    \caption{Debiasing for image captioning.}
    \label{fig:left}
  \end{subfigure}
  \begin{subfigure}[t]{0.45\textwidth}
    \centering
    \includegraphics[width=\textwidth]{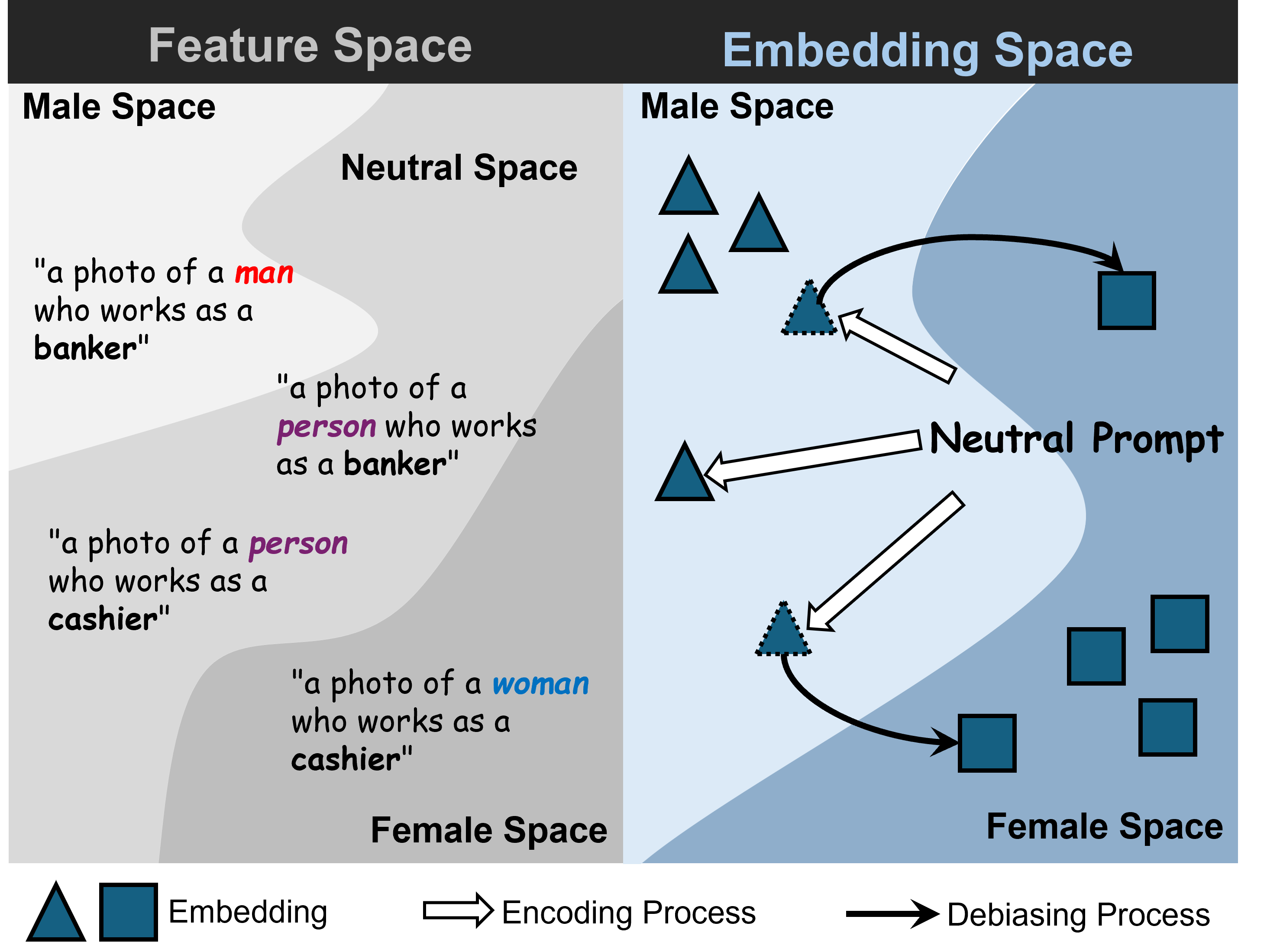}
    \caption{Debiasing for text-to-image generation.
}
    \label{fig:right}
  \end{subfigure}
  
\caption{
Overview of \sysname{} on two multimodal tasks. 
For image captioning, \sysname{} removes gender-related components by projecting representations onto the orthogonal complement of the gender subspace. 
For text-to-image generation, \sysname{} further introduces a calibration mechanism to balance gender distributions in generated images.
}
  \Description{Two-part pipeline diagram showing orthogonal projection for image captioning and projection with calibration for text-to-image generation.}
  \label{fig:method}
\end{figure*}
\subsection{Text-to-Image Generation}
For the image generation task, consider an input text prompt $t_g \in \mathcal{T}$ and a VLM $M_g$ that synthesizes an image $y_g = M_g(t_g)$. Similar to the captioning case, the prompt set $\mathcal{T}$ is partitioned into a {neutral set} $\mathcal{T}_n$, containing prompts without explicit gender specification, and an {explicit set} $\mathcal{T}_e$, which includes clear gender cues. Let $z(t_g)$ denote the original text embedding. The goal of difference-aware fairness is to learn a debiasing transformation $F_g$ such that the modified embedding $\tilde{z}(t_g) = F_g(z(t_g))$ simultaneously satisfies the following requirements:

\begin{equation}
\left\{
\begin{aligned}
    & \frac{\mathbb{P}\big(G(y_g) = \textit{male} \mid M_g(\tilde{z}(t_n))\big)}{\mathbb{P}\big(G(y_g) = \textit{female} \mid M_g(\tilde{z}(t_n))\big)} \approx 1,  \forall t_n \in \mathcal{T}_n \quad \text{(Neutral Fairness)} \\
    & \frac{\mathbb{P}\big(G(y_g) = g \mid M_g(\tilde{z}(t_e))\big)}{\mathbb{P}\big(G(y_g) = g \mid M_g(z(t_e))\big)} \approx 1,  \forall t_e \in \mathcal{T}_e \quad \text{(Explicit Faithfulness)} \\
    & d\big(\tilde{z}(t_g), z(t_g)\big) \le \epsilon_g,  \forall t_g \in \mathcal{T} \quad \text{(Semantic Preservation)}
\end{aligned}
\right.
\label{eq:t2i_fairness_goal}
\end{equation}

where $G(\cdot)$ is a gender classifier for generated images, $g \in \{\textit{male}, \textit{female}\}$ denotes the gender explicitly specified in $t_e$, and $\epsilon_g$ is a small tolerance constant. This formulation ensures that $F_g$ balances gender representation for neutral prompts while maintaining strict fidelity to the intended semantics when gender is explicitly defined.

\section{Methodology}

In this section, we introduce a novel debiasing method termed Bias Orthogonal Projection (\sysname{}). In Fig.\ref{fig:method}, we can see that the projection strategy adopted for the image captioning and image generation tasks differs slightly to accommodate the characteristics of each task.
Like Section \ref{sec: background}, subscripts $c$ and $g$ denote captioning and generation.

\subsection{Intuition and Task Differences}

As illustrated in Fig.~\ref{fig:method}, the behaviors of multimodal models differ across tasks, leading to distinct manifestations of gender bias.

For \textbf{image captioning}, both the feature space and the output embedding space can be conceptually decomposed into three subspaces: male, female, and neutral. 
Ideally, samples in the neutral feature subspace should be mapped to the neutral embedding subspace. 
However, biased models tend to misproject some neutral samples into gendered subspaces, resulting in incorrect gender descriptions. 
Our method corrects this issue by projecting representations back to the neutral subspace via orthogonal projection.

For \textbf{text-to-image generation}, the situation is different. 
Although the feature space still contains neutral semantics, the output space is inherently binary (male or female), since generated images must exhibit gender attributes. 
As a result, neutral prompts are often disproportionately mapped to one gender, leading to imbalanced outputs. 
To address this, we introduce a calibration mechanism that shifts part of the representations toward the underrepresented gender, improving balance.
\subsection{\sysname{} for Image Captioning}
\label{sec: biopro captioning}
\textbf{Construction of Gender-Variation Subspace.}
For a captioning model, given an input image $x$ and a prompt $t_c$, we obtain the fused hidden representation $h(x, t_c)$ and the corresponding matrix $\mathbf H$. Suppose we have two images that are similar in all aspects except for the gender attribute. Denote their embeddings as a pair $(\mathbf h_m, \mathbf h_f)$; the vector difference $\mathbf h_m - \mathbf h_f$ encodes high-purity gender-variation information. Specifically, we adopt the synthetic dataset \texttt{SCFs}~\cite{howard2024socialcounterfactuals}. \texttt{SCFs} leverages the method from~\cite{brooks2023instructpix2pix}, which injects cross-attention maps during the denoising steps to control the attention between certain pixels and tokens, thereby generating images that are similar in all features except for a clear difference in gender. At this point, we obtain the desired pairs $(\mathbf h_m, \mathbf h_f)$ and the corresponding matrices $\mathbf H_m, \mathbf H_f \in \mathbb{R}^{d \times n_1}$, where $n_1$ denotes the number of male and female samples in the synthetic dataset. In practice, we use 5,000 images for each gender. With the difference matrix $\mathbf D_c = \mathbf H_m - \mathbf H_f$, we perform Singular Value Decomposition (SVD) to capture the principal directions of gender variation:
\begin{equation}
    \mathbf D_c = \mathbf U_c \mathbf \Sigma_c \mathbf V_c^\top.
\end{equation}
The columns of $\mathbf U_c$ corresponding to the top $k$ singular values represent the most significant directions of gender variation. We take the first $k$ column vectors of $\mathbf U_c$ as the gender subspace:
\begin{equation}
    \mathbf S_c = \text{span}(\mathbf U_c[~:~, 1\!:\!k])=\text{span}(\mathbf U_c^k),
\end{equation}
where $\mathbf S_c \in \mathbb{R}^{d \times k}$ denotes the gender-variation subspace for captioning and $\mathbf U_c^k=\mathbf U_c[~:~, 1\!:\!k]$. 

\textbf{Orthogonal Projection for Debiasing.}
Once the gender-variation subspace $\mathbf S_c$ is obtained, we perform orthogonal projection to remove the gender-related components from the fused embeddings. 
Given the test embedding matrix $\mathbf H \in \mathbb{R}^{d \times n}$ to be debiased, we project it onto the orthogonal complement of $\mathbf S_c$ as follows:
\begin{equation}
    \mathbf H' = (\mathbf I - \mathbf U_c^k (\mathbf U_c^k)^\top) \mathbf H,
\end{equation}
where $\mathbf I \in \mathbb{R}^{d \times d}$ is the identity matrix. 
The matrix $\mathbf P_{\perp}=(\mathbf I - \mathbf U_c^k (\mathbf U_c^k)^\top)$ acts as an orthogonal projector that removes the components of $\mathbf H$ lying within the gender subspace $\mathbf S_c$, thus yielding the debiased representation $\mathbf H'$ for subsequent caption generation.
The orthogonal projection preserves task-relevant semantics by removing only gender-related components.
Let the fused representation $\mathbf H \in \mathbb{R}^{d \times n}$ be decomposed as
$\mathbf H = \mathbf H_{\text{bias}} + \mathbf H_{\text{sem}}$,
where $\mathbf H_{\text{bias}} \in \mathbf S_c$ denotes the gender-related component within the gender-variation subspace $\mathbf S_c$, and $\mathbf H_{\text{sem}} \in \mathbf S_c^\perp$ represents the semantic component in its orthogonal complement $\mathbf S_c^\perp$.
Applying the projection matrix $\mathbf P_{\perp}$ yields
$\mathbf H' = \mathbf \mathbf \mathbf P_{\perp}\mathbf H = \mathbf H_{\text{sem}}$.
Hence, the operation removes $\mathbf H_{\text{bias}}$ while preserving $\mathbf H_{\text{sem}}$, ensuring that semantic information (e.g., objects and actions) remains intact.
Since the dimension $k$ of $\mathbf S_c$ is small $(k \ll d)$, the impact on the overall representational capacity is minimal. Finally, $\mathbf H'$ is used to generate fair captions.
\begin{figure}[!t]
    \centering
    \includegraphics[width=\linewidth]{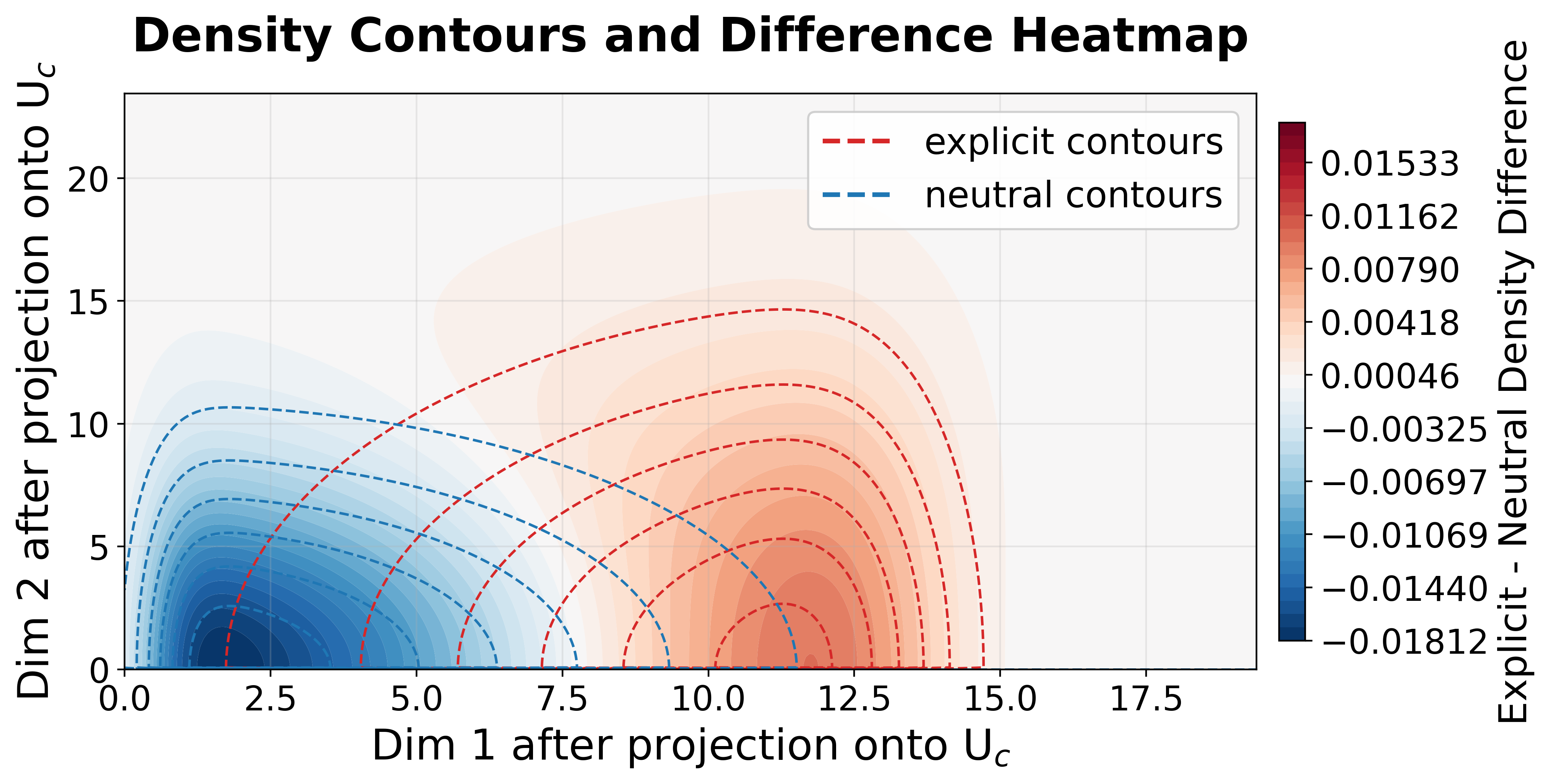}
    \caption{Absolute values after projection onto $\mathbf{U}_c$ when using the FLUX.1-dev base model.}
    \Description{Two-dimensional density contours comparing projected embedding magnitudes for neutral and explicitly gendered samples.}
    \label{fig: contour}
    \vspace{-6mm}
\end{figure}

\textbf{Projection-Based Selection.} Since the explicitly gendered image $x_e$ contains stronger bias information,  
the magnitude of its projection onto the  orthonormal basis $\mathbf U_c$ is larger than that of a neutral image $x_n$. Therefore, we can model the probability distribution of the absolute values of embeddings projected onto $\mathbf U_c$ using the validation set. As an example, we project the embeddings of the \texttt{MS-COCO}~\cite{lin2014microsoft} validation set onto the first two dimensions of $\mathbf U_c$ (see Fig.~\ref{fig: contour}). The explicit samples exhibit larger absolute values along both dimensions, with a more pronounced advantage in the first dimension. Furthermore, we model the projections of neutral and explicit samples on the first dimension as two skew-normal distributions, denoted as $p_n$ and $p_e$. We aim to find a threshold $\delta_c$, which approximately corresponds to the white boundary between the two probability density difference heatmaps in Fig.~\ref{fig: contour}. When the projection value of a sample is smaller than $\delta_c$, we apply the orthogonal projection; otherwise, we retain its original embedding. Therefore, the process of determining $\delta_c$ can be formulated as the following optimization problem:
\begin{equation}
\label{eq:delta_optimization}
\delta_c = \arg\max_{\delta} \left( 
    \int_{0}^{\delta} p_n \, dx 
    + \lambda_c \int_{\delta}^{+\infty} p_e \, dx
\right),
\end{equation}
where $\lambda_c$ is a trade-off coefficient that balances the preservation of explicit samples and the correction of potential bias in neutral ones. The first term encourages the projection to cover as many neutral samples as possible within the low-bias region, while the second term constrains the overlap with the explicit distribution to avoid over-correction. In practice, we approximate $\delta_c$ using Newton's method~\cite{nocedal2006numerical}.

\subsection{Calibration for Text-to-Image Generation}
For image generation, we can easily get gender variation space $\mathbf S_g$ and compute $\mathbf P_{\perp}$ by constructing prompt pairs such as ``\textit{a photo of a male person who works as a doctor}'' and ``\textit{a photo of a female person who works as a doctor}''.
However, the image generation and captioning tasks differ in their output spaces (Fig.~\ref{fig:right}).  
For the \text{image captioning} task, the model can generate gendered words to describe the person in the image,  
as well as neutral terms such as ``\textit{a person}''.  
Our orthogonal projection increases the probability of generating such neutral expressions.  
However, for the \text{image generation} task, the generated image of a person necessarily contains gender attributes.  
After projection, while the probability of generating a biased gender may be reduced,  
it can still remain higher than that of the opposite gender,  
resulting in suboptimal debiasing performance.  
To address this issue, we introduce a \text{calibration term} to improve the balance. 
For example, if a neutral prompt tends to generate female images with an extreme bias, we formulate the following optimization objective:
\begin{equation}
    \min_{\mathbf P} 
        \underbrace{\| \mathbf P - \mathbf P_{\perp} \|^2}_{\textbf{(i) Orthogonal Term}}
        + 
        \lambda_g \underbrace{\| \mathbf P \mathbf Z_f - \mathbf Z_m \|^2}_{\textbf{(ii) Calibration Term}},
        \label{eq: calibration}
\end{equation}
where $\mathbf P \in \mathbb{R}^{d \times d}$ denotes the learnable projection matrix, and $\mathbf Z_m$ and $\mathbf Z_f$ denote the latent embeddings corresponding to male and female prompts respectively. 
In this formulation, the \textbf{(i) Orthogonal Term} provides a good initialization for $\mathbf P$,  
constraining it to optimize along directions orthogonal to the gender subspace.  
This term ensures that $\mathbf P$ approximately preserves the orthogonal property of $\mathbf P_{\perp}$, 
preventing the embedding space from being excessively stretched or collapsed,  
which could otherwise lead to the loss of semantic information.  
The \textbf{(ii) Calibration Term} encourages $\mathbf P$ to project female representations toward the male embedding space,  
thereby calibrating the generative direction of the model.  
This adjustment helps balance the gender distribution of generated images and mitigates residual gender bias. 
\begin{lemma}[Closed-Form Solution for Calibration]
The optimization problem defined in the objective~\ref{eq: calibration} admits a closed-form solution.  
Specifically, the optimal projection matrix $\mathbf P_{f \to m}$ that minimizes the objective \ref{eq: calibration}
is given by
\begin{equation}
    \mathbf P_{f \to m} = 
    \big( \mathbf P_{\perp} + \lambda_g \mathbf Z_m \mathbf Z_f^{\top} \big)
    \big( \mathbf I + \lambda_g \mathbf Z_f \mathbf Z_f^{\top} \big)^{-1}.
    \label{eq: closed_form_P}
\end{equation}
\label{lemma: solution}

\end{lemma}

Furthermore, by obtaining the matrix $\mathbf P_{m \to f}$ corresponding to Eq.~\ref{eq: closed_form_P},  
we can effectively perform debiased image generation using these two projection matrices.

\begin{table*}[!t] 

\centering 
\caption{Evaluation results for image captioning. 
\textbf{Bold} indicates the best result, while \underline{underline} denotes the second-best result.} 
\label{tab: captioning}
\setlength{\tabcolsep}{3pt} 
\renewcommand{\arraystretch}{1}
\begin{tabular}{lcccccccccc} 
\toprule 
\multicolumn{1}{l}{\multirow{2}{*}{\textbf{Method}}} & 
\multicolumn{5}{c}{\textbf{LLaVA-1.5}} & 
\multicolumn{5}{c}{\textbf{LLaVA-NeXT}} \\ 
\cmidrule(lr){2-6} \cmidrule(lr){7-11}
&$\mathrm{BR}_n$$\downarrow$  &$\mathrm{BR}_e$$\rightarrow$$\mathrm{BR}_e^\text{base}$   &$\mathrm{CBR}$$\downarrow$ &METEOR$\uparrow$  &CLIP-S$\uparrow$   
&$\mathrm{BR}_n$$\downarrow$ &$\mathrm{BR}_e$$\rightarrow$$\mathrm{BR}_e^\text{base}$&$\mathrm{CBR}$$\downarrow$ &METEOR$\uparrow$  &CLIP-S$\uparrow$ \\ 
\midrule
Base &36.22  &80.27 &36.22 &0.317  &0.316  &15.87  &72.55 &\underline{15.87} &0.237  &0.330 \\

Prompt-1 &21.83  &61.83  &28.58  &0.325  &0.316  &8.32  &48.89  &25.08 &0.236  &0.329 \\ 
Prompt-2 &16.34  &54.92  &30.16  &0.324  &0.315  &7.08  &38.69  &34.59 &0.235  &0.330 \\ 
LIBRA~\cite{hirota2023model} &64.13  &90.19  &64.89  &0.339  &0.309  &68.97  &93.42  &72.06 &0.263 &0.314 \\ 
SFID~\cite{jung2024unified} &35.46  &79.72  &35.46  &0.317  &0.316  &16.34  &73.38  &16.36  &0.238  &0.330 \\ 
\midrule
\rowcolor{gray!15}\sysname{} &23.01  &68.74  &\textbf{25.74} &0.315  &0.315 &12.27  &64.06 &\textbf{14.92} &0.238  &0.329  \\ 
w/o Selection &20.29  &61.92  &\underline{27.36} &0.312  &0.314 &11.33  &55.52 &20.45 &0.236  &0.328  \\ 
\bottomrule 
\end{tabular}
\end{table*}

\begin{table}[t]
\centering

\caption{Sensitivity Analysis under different $(k, \lambda_c)$ configurations for LLaVA-1.5.}
\label{tab:sensitivity-llava15}
\setlength{\tabcolsep}{2pt}
\renewcommand{\arraystretch}{1.1}
\begin{tabular}{lccccc}
\toprule
\multicolumn{6}{c}{\textbf{LLaVA-1.5}} \\
\cmidrule(lr){2-6}
\multicolumn{1}{l}{\textbf{$(k,\lambda_c)$}} 
& $\mathrm{BR}_n\!\downarrow$ 
& $\mathrm{BR}_e \!\rightarrow\! \mathrm{BR}_e^\text{base}$ 
& $\mathrm{CBR}\!\downarrow$ 
& METEOR$\uparrow$  
& CLIP-S$\uparrow$ \\
\midrule
(1, 2) & 28.44 & 74.09 & 29.10 & 0.316 & 0.316 \\
(1, 3) & 26.78 & 72.86 & 27.79 & 0.316 & 0.316 \\
(1, 4) & 25.90 & 71.92 & 27.21 & 0.316 & 0.316 \\
(2, 2) & 25.19 & 70.34 & 27.07 & 0.316 & 0.315 \\
(2, 3) & 23.01 & 68.74 & 25.74 & 0.315 & 0.315 \\
(2, 4) & 21.53 & 67.31 & 25.13 & 0.314 & 0.315 \\
(4, 2) & 26.02 & 70.88 & 27.66 & 0.315 & 0.315 \\
(4, 3) & 24.01 & 69.35 & 26.37 & 0.315 & 0.315 \\
(4, 4) & 22.36 & 68.00 & 25.51 & 0.314 & 0.315 \\
\bottomrule
\end{tabular}

\end{table}

\section{Experiments on Image Captioning}
\label{sec: captioning experiments}

\subsection{Settings}

\textbf{Datasets.}
For image captioning, we sample 5,000 male-female image pairs from \texttt{SCFs}~\cite{howard2024socialcounterfactuals} to compute $\mathbf P_\perp$,
and use the gender-annotated \texttt{MS-COCO}~\cite{zhao2021understanding} for evaluation. \texttt{MS-COCO} contains 3,410 images labeled ``unsure gender'' due to visual ambiguity (e.g., occlusion or view from behind), and 10,780 images with explicit gender labels. In practice, we divide \texttt{MS-COCO} into validation and test sets with a 1:1 ratio.

\textbf{Baselines.} 
We employ LLaVA-v1.5-7b-hf (LLaVA-1.5)~\cite{liu2024improved} and LLaVA-v1.6-mistral-7b-hf  (LLaVA-NeXT)~\cite{liu2024llavanext} as the backbone models for image captioning. Meanwhile, we adopt Prompting, LIBRA~\cite{hirota2023model}, and SFID~\cite{jung2024unified} as comparison baselines.  
Specifically, Prompting refers to instructing the model via textual prompts, not to assign a gendered subject when the input image exhibits gender ambiguity.

\textbf{Metrics.} We define the proportion of captions containing gendered words among all generated captions as the bias rate ($\mathrm{BR}$).  
For neutral and explicit gender samples, we denote the bias rates as $\mathrm{BR}_n$ and $ \mathrm{BR}_e$, respectively.  
Our objective is to maintain a low $\mathrm{BR}_n$ while keeping $ \mathrm{BR}_e$ as close as possible to that of the base model.
Similar to the \text{composite misclassification rate}~\cite{jung2024unified}, we propose the \text{composite bias rate} ($\mathrm{CBR}$), defined as \begin{equation}
    \mathrm{CBR} = \sqrt{\mathrm{BR}_n^2 + (\mathrm{BR}_e - \mathrm{BR}_e^{\text{base}})^2},
\end{equation}
where $\mathrm{BR}_e^{\text{base}}$ denote the $\mathrm{BR}_e$ value of the base model.
$\mathrm{CBR}$ is designed to comprehensively evaluate both \text{neutral fairness} and \text{explicit gender faithfulness}.
To evaluate the semantic preservation of the generated captions, we adopt the METEOR~\cite{banerjee2005meteor} and CLIP Score (CLIP-S)~\cite{hessel2021clipscore} metrics, both of which are higher-is-better indicators.

\subsection{Overall Performance}

\begin{table*}[!t]
\centering
\caption{Evaluation results for image generation. \textbf{Bold} indicates the best result, while \underline{underline} denotes the second-best result. \textbf{MUSIQ} refers to the MUSIQ-KonIQ technical quality score (higher is better).}
\label{tab: generation}
\setlength{\tabcolsep}{3pt} 
\renewcommand{\arraystretch}{1}
\begin{tabular}{lccccccccccccc}
\toprule
\multirow{2}{*}{\textbf{Method}} &
\multicolumn{6}{c}{\textbf{FLUX.1-dev}} &
\multicolumn{6}{c}{\textbf{FLUX.1-schnell}} \\
\cmidrule(lr){2-7} \cmidrule(lr){8-13}
 & $Skew_m$$\downarrow$ & $Skew_f$$\downarrow$ & $Skew$$\downarrow$ & $\mathrm{MR}$$\downarrow$ & CLIP-S$\uparrow$ & MUSIQ$\uparrow$ 
 & $Skew_m$$\downarrow$ & $Skew_f$$\downarrow$ & $Skew$$\downarrow$ & $\mathrm{MR}$$\downarrow$ & CLIP-S$\uparrow$ & MUSIQ$\uparrow$ \\
\midrule
Base & 89.6 & 96.8 & 93.2 & 0 & 0.291 & 75.42 & 98.4 & 98.6 & 98.5 & 0 & 0.291 & 76.85 \\
BendVLM~\cite{gerych2024bendvlm} & 80.8 & 96.4 & 88.6 & 0 & 0.286 & 74.38 & 89.2 & 97.6 & 93.4 & 0.1\% & 0.285 & 73.53 \\
SFID~\cite{jung2024unified} & 93.0 & 94.6 & 93.8 & 0 & 0.289 & 75.56 & 99.2 & 98.8 & 99.0 & 0 & 0.291 & 77.09 \\
Prompt-Projection~\cite{chuang2023debiasing} & 94.2 & 90.4 & 92.3 & 0 & 0.289 & 75.58 & 99.4 & 98.0 & 98.7 & 0 & 0.290 & 77.03 \\
ForcePrompt~\cite{fu2025fairimagen} & 75.4 & 99.0 & 87.2 & 0 & \underline{0.294} & 75.81 & 78.2 & 100.0 & 89.1 & 0 & \underline{0.295} & 77.28 \\
FairImagen~\cite{fu2025fairimagen} & \underline{68.0} & \underline{77.0} & \underline{72.5} & 0 & 0.291 & 75.41 & \underline{60.0} & \underline{69.2} & \underline{64.6} & 0 & 0.291 & 76.84 \\
\midrule
\rowcolor{gray!15}
\sysname{} & \textbf{60.8} & \textbf{74.8} & \textbf{67.8} & 0.2\% & 0.288 & 75.84 & \textbf{58.6} & \textbf{62.2} & \textbf{60.4} & 0.2\% & 0.290 & 76.69 \\
w/o Calibration & 87.8 & 96.4 & 92.1 & 0 & 0.290 & 75.88 & 97.8 & 96.4 & 97.1 & 0 & 0.291 & 77.02 \\
\bottomrule
\end{tabular}
\end{table*}
As shown in Table~\ref{tab: captioning}, we present the overall captioning results. 
Lower values of $\mathrm{BR}_n$ and $\mathrm{CBR}$ indicate better performance, while $\mathrm{BR}_e$ is preferable when it is closer to its baseline $\mathrm{BR}_e^\text{base}$. 
For the semantic metrics, both METEOR and CLIP-S are the higher, the better. In both \text{LLaVA-1.5} and \text{LLaVA-NeXT}, \sysname{} achieves the best $\mathrm{CBR}$, indicating that our method can effectively reduce the model’s biased summarization on images with unknown gender while avoiding excessive intervention in the captions of images with explicit gender. It is worth noting that although directly modifying the prompt may affect user experience, such methods can effectively mitigate bias. However, the model is sensitive to different prompts, and for samples with explicit gender, prompt-based methods cannot guarantee \text{explicit gender faithfulness}. On \text{LLaVA-NeXT}, \sysname{} is the only model that achieves a positive improvement in $\mathrm{CBR}$. In terms of the quality of generated captions, both \text{METEOR} and \text{CLIP-S} remain largely consistent with those of the original model across all methods, indicating that \sysname{} ensures \text{semantic preservation}.

\textbf{Ablation Study (\text{w/o Selection}).}
We remove the projection-based selection module to examine its effect in $\mathrm{BR}_n$ and $\mathrm{BR}_e$. We can observe that although further gains are achieved on $\mathrm{BR}_n$, the loss on $\mathrm{BR}_e$ is significant, resulting in a larger deviation from $\mathrm{BR}_e^\text{base}$, and thus failing to achieve the optimal $\mathrm{CBR}$. This setting corresponds to a non-selective, global debiasing approach.

\subsection{Sensitivity Analysis}
\label{app:sensitivity}

We analyze the impact of the gender-variation subspace dimension $k$ and the selection coefficient $\lambda_c$ on the image captioning task. Table~\ref{tab:sensitivity-llava15} presents the results on LLaVA-1.5. We report $\mathrm{BR}_n$, $\mathrm{BR}_e$, $\mathrm{CBR}$, and semantic preservation metrics (METEOR and CLIP-S).

Specifically, we vary the subspace dimension $k \in \{1, 2, 4\}$ to control the capacity of the extracted bias subspace, and adjust the selection coefficient $\lambda_c \in \{2, 3, 4\}$ to modulate the strength of selective debiasing.
Table~\ref{tab:sensitivity-llava15} indicates that \sysname{} maintains stable performance across this range of parameter settings, effectively balancing fairness and semantic fidelity.

\section{Experiments on Text-to-Image Generation}
\label{sec: generation experiments}

\subsection{Settings}
\textbf{Prompt Construction.} Following  \cite{jung2024unified}, we utilize the input prompt from \cite{cho2023dall} to generate images for captions: ``\textit{a photo of $\mathcal{G}$ who works as a/an $\mathcal{P}$}", where gender set $\mathcal{G}$ = \{\textit{a man}, \textit{a woman}, \textit{a person}\} and profession set $\mathcal{P}$ = \{\textit{banker}, \textit{teacher}, \textit{...}\}. We select five male-stereotyped and five female-stereotyped occupations to construct the test prompts, and 90 occupations to construct the training prompts.  
In particular, we employ \text{GPT-5}~\cite{singh2025openai} to assist in generating diverse prompt templates for producing additional training prompts to calculate a better $\mathbf P_{\perp}$.

\textbf{Baselines.} We employ FLUX.1-dev\footnote{https://huggingface.co/black-forest-labs/FLUX.1-dev} and FLUX.1-schnell\footnote{https://huggingface.co/black-forest-labs/FLUX.1-schnell} as the backbone models for image generation. FLUX is a next-generation generative model based on the \text{flow matching}~\cite{lipman2022flow} framework.  We compare two categories of debiasing methods. The first category requires real images as auxiliary inputs for bias mitigation, including \text{BendVLM}~\cite{gerych2024bendvlm} and  \text{SFID}~\cite{jung2024unified}. The second category does not rely on additional images, which includes \text{Prompt-Projection}~\cite{chuang2023debiasing}, ForcePrompt~\cite{fu2025fairimagen}, \text{FairImagen}~\cite{fu2025fairimagen} and our proposed method: \sysname{}. For a fair comparison, each method generates 100 images per prompt with fixed seeds (0–99).

\textbf{Metrics.} We adopt $Skew$~\cite{jung2024unified} as the metric to evaluate neutral fairness, which can be computed as follows:
\begin{equation}
    Skew=\frac{1}{|\mathcal{P}|} \sum_{p \in \mathcal{P}} \frac{\max \left(N_{p, m}, N_{p, f}\right)}{C},
\end{equation}
where $N_{p, m}$ and $N_{p, f}$ are the numbers of detected genders for each profession, and $C = 100$ is the number of generation for each prompt. We use BLIP-2~\cite{li2023blip} to detect the gender of generated images by asking ``\textit{Does the person look like a male or a female?}"~\cite{jung2024unified}. In addition to the overall $Skew$, we also measure $Skew_m$ and $Skew_f$ for male-stereotyped and female-stereotyped professions, respectively. To evaluate explicit gender faithfulness for explicit prompts, we calculate the misclassification rate ($\mathrm{MR}$) of all the images. We do not expect the reduction of $Skew$ to excessively increase the $\mathrm{MR}$.
Similar to captioning, we use the \text{CLIP-S} to evaluate semantic preservation. Furthermore, we adopt MUSIQ~\cite{ke2021musiq} to evaluate the quality of the generated images.

\subsection{Overall Performance}
From the image generation results shown in Table~\ref{tab: generation}, we can observe that, unlike the image captioning task, all methods maintain \text{explicit gender faithfulness} in the image generation task. Only the two methods that achieve successful debiasing performance (BendVLM and \sysname{}) exhibit a very slight error rate (0.1\% and 0.2\%). Therefore, to simplify the pipeline, we do not apply the projection-based selection for selective debiasing in generation task. From the perspective of overall debiasing performance, \sysname{} achieves the best results on both gender–specific $Skew$ and overall $Skew$, while introducing only a negligible error rate. It maintains a CLIP-S nearly identical to the base model, with no drop in MUSIQ.
In addition, the advantage of \sysname{} lies in its generalizable paradigm, applicable to both image captioning tasks (Section~\ref{sec: captioning experiments}) and continuous bias variables (Section~\ref{sec: scene bias}).

\textbf{Ablation Study.}  
In Table~\ref{tab: generation}, we present a variant of \sysname{}, where only a single term (i) is retained in Eq.~\ref{eq: calibration}. \textbf{W/o Calibration}: It can be observed that the debiasing effect weakens considerably, while $\mathrm{MR}$ drops to zero, which further reveals that debiasing may cause the model to generate a small portion of incorrect outputs.
\section{Debiasing Continuous Variables: Scene Bias}
\begin{figure*}[!t]
\centering
    \includegraphics[width=0.9\linewidth]{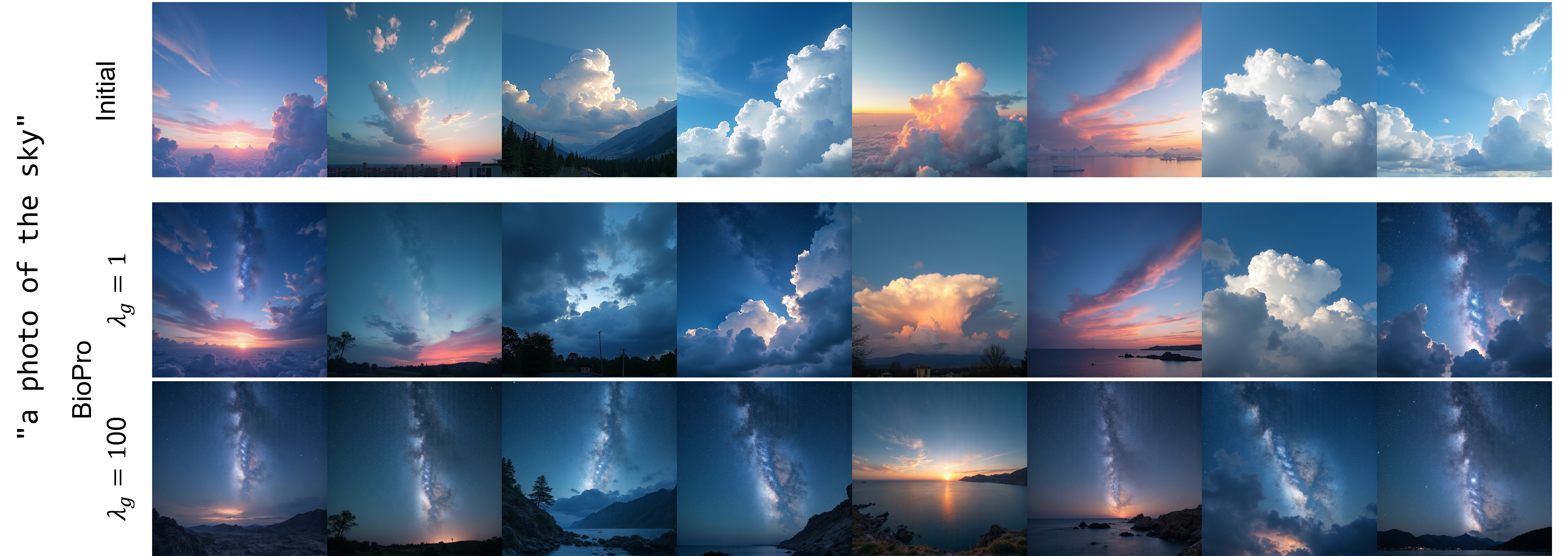}
    \vspace{-2mm}
    \caption{Debiasing scene bias with \sysname{}. The figure shows images generated with fixed seeds ranging from
0 to 7.}
    \Description{Eight generated sky images showing controllable changes from bright daytime scenes toward darker nighttime scenes.}
    \label{fig: sky}
\end{figure*}
\label{sec: scene bias}

We observe that in text-to-image generation tasks, bias does not only occur in discrete variables such as gender, but also in continuous variables such as scene brightness. For example, when the lighting condition is unspecified, and the model is prompted to generate a photo of the sky, it tends to produce images under bright lighting conditions. In contrast, \sysname{} is capable of mitigating such continuous bias variables. Given a neutral prompt like ``\textit{a photo of the sky}", it can adjust the brightness level of the generated image to produce darker skies, where the degree of darkness can be controlled by the parameter $\lambda_g$ in Eq.~\ref{eq: calibration}.
Like gender bias, we generate images for captions: ``\textit{a photo of the $\mathcal{O}$}”, where object set $\mathcal{O}$ = \{\text{sky}, \text{forest}, \text{grassland} ,\text{sea}\}.
We obtain $\mathbf P_{\perp}$ by constructing prompt pairs such as ``\textit{a light photo of the river}" and ``\textit{a dark photo of the river}".
Similarly, we utilize prompt templates to produce additional training prompts for calculating a better $\mathbf P_{\perp}$, and generate images using seeds 0--99 for each neutral prompt.
Fig.~\ref{fig: sky} shows the generated images of the sky. \sysname{} successfully adjusts the brightness of the generated images, and by increasing $\lambda_g$, we can enhance the darkness level to produce more nighttime skies. 

\textbf{Metrics.}
We quantify bias mitigation with task-specific metrics; e.g., for brightness we directly measure luminance statistics (Lum.)~\cite{reinhard2002color}. To evaluate the quality of the generated images, we also employed CLIP-S and MUSIQ.

\textbf{Overall Performance.}
Table~\ref{tab: scene bias} summarizes the quantitative results on scene bias debiasing.
We observe that \sysname{} consistently preserves generation quality while effectively
mitigating brightness bias. Specifically, the CLIP-S and MUSIQ remain nearly identical
before and after debiasing across both FLUX.1-dev and FLUX.1-schnell, indicating that
our method introduces negligible semantic distortion. This demonstrates that the proposed
projection-based intervention operates in a highly targeted manner, removing bias-related
components without degrading the alignment between generated images and input prompts.
Beyond semantic fidelity, \sysname{} improves the diversity of generated outputs along
the continuous brightness dimension. While the base model exhibits a strong tendency
toward overexposed or brightly lit scenes under neutral prompts, our method enables a
adjustable luminance distribution, as reflected by the controlled adjustment of Lum.
statistics. Importantly, this adjustment is not achieved through stochastic variation,
but through a deterministic and controllable mechanism governed by the parameter
$\lambda_g$, which allows smooth interpolation between brighter and darker visual modes.
\begin{table}[!t]
\small
\centering
\caption{Results of debiasing scene bias.}
\label{tab: scene bias}
\setlength{\tabcolsep}{3.5pt}
\renewcommand{\arraystretch}{1.1}
\begin{tabular}{lcccccc}
\toprule
\multirow{2}{*}{\textbf{Method}} & \multicolumn{3}{c}{\textbf{FLUX.1-dev}} & \multicolumn{3}{c}{\textbf{FLUX.1-schnell}} \\
\cmidrule(lr){2-4} \cmidrule(lr){5-7}
 & CLIP-S $\uparrow$ & MUSIQ $\uparrow$ & Lum.  & CLIP-S $\uparrow$ & MUSIQ $\uparrow$ & Lum. \\
\midrule
Base & 0.272 & 64.59 & 51.23 &0.278  & 54.00 & 52.68 \\
\sysname{} & 0.274 & 63.16 & 42.36 &0.278  & 50.42 & 31.42 \\
\bottomrule
\end{tabular}
\end{table}

Furthermore, compared to discrete bias mitigation (e.g., gender), debiasing continuous
variables poses additional challenges due to the absence of clear categorical boundaries.
The strong performance of \sysname{} in this setting highlights its ability to capture
fine-grained bias directions in the representation space and perform calibrated
manipulation along these directions. This suggests that our framework generalizes beyond
binary or categorical attributes, providing a unified solution for both discrete and
continuous bias mitigation.
Overall, the results demonstrate that \sysname{} not only maintains high visual and
semantic quality, but also significantly enriches the controllability and diversity of
text-to-image generation, making it a practical and flexible tool for bias-aware
generation.

\section{Discussion}
While our current formulation primarily focuses on binary attributes (e.g., gender),
the proposed framework naturally extends to multi-class discrete variables.
Instead of constructing a single bias direction, we can estimate multiple
attribute-specific directions by forming contrastive prompt pairs across all
categories.
These directions can then be organized into a subspace that captures the full
spectrum of attribute variations. From a representation learning perspective, our method can be interpreted
as a linear intervention in the latent space. While this simplicity enables
efficient and training-free deployment, it may not fully account for nonlinear
entanglement between attributes. In highly entangled scenarios, bias-related and
task-relevant features may not be perfectly separable via linear projection,
leading to a trade-off between debiasing strength and content preservation.


\section{Conclusion}

In this work, we extend the principle of context-sensitive fairness from
text-only models to multimodal generation. We propose \sysname{}, a
training-free debiasing framework that performs selective bias mitigation
through orthogonal projection in the representation space. By identifying
and removing bias-related components while preserving task-relevant
information, our method achieves a strong balance between fairness and
generation quality.
Beyond conventional categorical bias, we further demonstrate that \sysname{}
generalizes to continuous bias variables, such as scene brightness, enabling
fine-grained and controllable manipulation of generated content. 
This
Overall, our work provides a controllable and effective paradigm for bias-aware
generation.

\section*{Acknowledgements}
The project was supported by National Key\&D Program of China (No. 2022ZD0160501), Natural Science Foundation of Fujian Province of China (No. 2024J011001), and the Public Technology Service
Platform Project of Xiamen (No.3502Z20231043).
We also thank the reviewers for their insightful comments.
\bibliographystyle{ACM-Reference-Format}
\bibliography{reference}

@String(ICCV= {Int. Conf. Comput. Vis.})

@String(AAAI = {AAAI})

@String(ICCV  = {ICCV})

@inproceedings{howard2024socialcounterfactuals,
  title={Socialcounterfactuals: Probing and mitigating intersectional social biases in vision-language models with counterfactual examples},
  author={Howard, Phillip and Madasu, Avinash and Le, Tiep and Moreno, Gustavo Lujan and Bhiwandiwalla, Anahita and Lal, Vasudev},
  booktitle={Proceedings of the IEEE/CVF Conference on Computer Vision and Pattern Recognition},
  pages={11975--11985},
  year={2024}
}

@inproceedings{brooks2023instructpix2pix,
  title={Instructpix2pix: Learning to follow image editing instructions},
  author={Brooks, Tim and Holynski, Aleksander and Efros, Alexei A},
  booktitle={Proceedings of the IEEE/CVF conference on computer vision and pattern recognition},
  pages={18392--18402},
  year={2023}
}

@inproceedings{lin2014microsoft,
  title={Microsoft coco: Common objects in context},
  author={Lin, Tsung-Yi and Maire, Michael and Belongie, Serge and Hays, James and Perona, Pietro and Ramanan, Deva and Doll{\'a}r, Piotr and Zitnick, C Lawrence},
  booktitle={European conference on computer vision},
  pages={740--755},
  year={2014},
  organization={Springer}
}

@inproceedings{zhao2021understanding,
  title={Understanding and Evaluating Racial Biases in Image Captioning},
  author={Zhao, Dora and Wang, Angelina and Russakovsky, Olga},
  booktitle=ICCV,
  pages={14830--14840},
  year={2021}
}

@article{wang2025fairness,
  title={Fairness through difference awareness: Measuring desired group discrimination in LLMs},
  author={Wang, Angelina and Phan, Michelle and Ho, Daniel E and Koyejo, Sanmi},
  journal={arXiv preprint arXiv:2502.01926},
  year={2025}
}

@article{achiam2023gpt,
  title={Gpt-4 technical report},
  author={Achiam, Josh and Adler, Steven and Agarwal, Sandhini and Ahmad, Lama and Akkaya, Ilge and Aleman, Florencia Leoni and Almeida, Diogo and Altenschmidt, Janko and Altman, Sam and Anadkat, Shyamal and others},
  journal={arXiv preprint arXiv:2303.08774},
  year={2023}
}

@article{dubey2024llama,
  title={The llama 3 herd of models},
  author={Dubey, Abhimanyu and Jauhri, Abhinav and Pandey, Abhinav and Kadian, Abhishek and Al-Dahle, Ahmad and Letman, Aiesha and Mathur, Akhil and Schelten, Alan and Yang, Amy and Fan, Angela and others},
  journal={arXiv e-prints},
  pages={arXiv--2407},
  year={2024}
}

@inproceedings{liang2021towards,
  title={Towards understanding and mitigating social biases in language models},
  author={Liang, Paul Pu and Wu, Chiyu and Morency, Louis-Philippe and Salakhutdinov, Ruslan},
  booktitle={International conference on machine learning},
  pages={6565--6576},
  year={2021},
  organization={PMLR}
}

@article{parrish2021bbq,
  title={BBQ: A hand-built bias benchmark for question answering},
  author={Parrish, Alicia and Chen, Angelica and Nangia, Nikita and Padmakumar, Vishakh and Phang, Jason and Thompson, Jana and Htut, Phu Mon and Bowman, Samuel R},
  journal={arXiv preprint arXiv:2110.08193},
  year={2021}
}

@article{gallegos2024bias,
  title={Bias and fairness in large language models: A survey},
  author={Gallegos, Isabel O and Rossi, Ryan A and Barrow, Joe and Tanjim, Md Mehrab and Kim, Sungchul and Dernoncourt, Franck and Yu, Tong and Zhang, Ruiyi and Ahmed, Nesreen K},
  journal={Computational Linguistics},
  volume={50},
  number={3},
  pages={1097--1179},
  year={2024},
  publisher={MIT Press 255 Main Street, 9th Floor, Cambridge, Massachusetts 02142, USA~…}
}

@article{nadeem2020stereoset,
  title={StereoSet: Measuring stereotypical bias in pretrained language models},
  author={Nadeem, Moin and Bethke, Anna and Reddy, Siva},
  journal={arXiv preprint arXiv:2004.09456},
  year={2020}
}

@article{liu2023visual,
  title={Visual instruction tuning},
  author={Liu, Haotian and Li, Chunyuan and Wu, Qingyang and Lee, Yong Jae},
  journal={Advances in neural information processing systems},
  volume={36},
  pages={34892--34916},
  year={2023}
}

@article{zhu2023minigpt,
  title={Minigpt-4: Enhancing vision-language understanding with advanced large language models},
  author={Zhu, Deyao and Chen, Jun and Shen, Xiaoqian and Li, Xiang and Elhoseiny, Mohamed},
  journal={arXiv preprint arXiv:2304.10592},
  year={2023}
}

@inproceedings{rombach2022high,
  title={High-resolution image synthesis with latent diffusion models},
  author={Rombach, Robin and Blattmann, Andreas and Lorenz, Dominik and Esser, Patrick and Ommer, Bj{\"o}rn},
  booktitle={Proceedings of the IEEE/CVF conference on computer vision and pattern recognition},
  pages={10684--10695},
  year={2022}
}

@inproceedings{liu2024improved,
  title={Improved baselines with visual instruction tuning},
  author={Liu, Haotian and Li, Chunyuan and Li, Yuheng and Lee, Yong Jae},
  booktitle={Proceedings of the IEEE/CVF conference on computer vision and pattern recognition},
  pages={26296--26306},
  year={2024}
}

@misc{liu2024llavanext,
  title={Llavanext: Improved reasoning, ocr, and world knowledge},
  author={Liu, Haotian and Li, Chunyuan and Li, Yuheng and Li, Bo and Zhang, Yuanhan and Shen, Sheng and Lee, Yong Jae},
  year={2024}
}

@article{jung2024unified,
  title={A unified debiasing approach for vision-language models across modalities and tasks},
  author={Jung, Hoin and Jang, Taeuk and Wang, Xiaoqian},
  journal={Advances in Neural Information Processing Systems},
  volume={37},
  pages={21034--21058},
  year={2024}
}

@inproceedings{hirota2023model,
  title={Model-agnostic gender debiased image captioning},
  author={Hirota, Yusuke and Nakashima, Yuta and Garcia, Noa},
  booktitle={Proceedings of the IEEE/CVF Conference on Computer Vision and Pattern Recognition},
  pages={15191--15200},
  year={2023}
}

@inproceedings{banerjee2005meteor,
  title={METEOR: An automatic metric for MT evaluation with improved correlation with human judgments},
  author={Banerjee, Satanjeev and Lavie, Alon},
  booktitle={Proceedings of the acl workshop on intrinsic and extrinsic evaluation measures for machine translation and/or summarization},
  pages={65--72},
  year={2005}
}

@article{hessel2021clipscore,
  title={Clipscore: A reference-free evaluation metric for image captioning},
  author={Hessel, Jack and Holtzman, Ari and Forbes, Maxwell and Bras, Ronan Le and Choi, Yejin},
  journal={arXiv preprint arXiv:2104.08718},
  year={2021}
}

@inproceedings{cho2023dall,
  title={Dall-eval: Probing the reasoning skills and social biases of text-to-image generation models},
  author={Cho, Jaemin and Zala, Abhay and Bansal, Mohit},
  booktitle={Proceedings of the IEEE/CVF international conference on computer vision},
  pages={3043--3054},
  year={2023}
}

@article{shao2024supervised,
  title={Supervised algorithmic fairness in distribution shifts: A survey},
  author={Shao, Minglai and Li, Dong and Zhao, Chen and Wu, Xintao and Lin, Yujie and Tian, Qin},
  journal={arXiv preprint arXiv:2402.01327},
  year={2024}
}

@book{nocedal2006numerical,
  title={Numerical optimization},
  author={Nocedal, Jorge and Wright, Stephen J},
  year={2006},
  publisher={Springer}
}

@article{singh2025openai,
  title={Openai gpt-5 system card},
  author={Singh, Aaditya and Fry, Adam and Perelman, Adam and Tart, Adam and Ganesh, Adi and El-Kishky, Ahmed and McLaughlin, Aidan and Low, Aiden and Ostrow, AJ and Ananthram, Akhila and others},
  journal={arXiv preprint arXiv:2601.03267},
  year={2025}
}

@article{lipman2022flow,
  title={Flow matching for generative modeling},
  author={Lipman, Yaron and Chen, Ricky TQ and Ben-Hamu, Heli and Nickel, Maximilian and Le, Matt},
  journal={arXiv preprint arXiv:2210.02747},
  year={2022}
}

@article{gerych2024bendvlm,
  title={Bendvlm: Test-time debiasing of vision-language embeddings},
  author={Gerych, Walter and Zhang, Haoran and Hamidieh, Kimia and Pan, Eileen and Sharma, Maanas K and Hartvigsen, Tom and Ghassemi, Marzyeh},
  journal={Advances in Neural Information Processing Systems},
  volume={37},
  pages={62480--62502},
  year={2024}
}

@article{fu2025fairimagen,
  title={FairImagen: Post-Processing for Bias Mitigation in Text-to-Image Models},
  author={Fu, Zihao and Brown, Ryan and Shao, Shun and Rawal, Kai and Delaney, Eoin and Russell, Chris},
  journal={arXiv preprint arXiv:2510.21363},
  year={2025}
}

@article{chuang2023debiasing,
  title={Debiasing vision-language models via biased prompts},
  author={Chuang, Ching-Yao and Jampani, Varun and Li, Yuanzhen and Torralba, Antonio and Jegelka, Stefanie},
  journal={arXiv preprint arXiv:2302.00070},
  year={2023}
}

@inproceedings{li2023blip,
  title={Blip-2: Bootstrapping language-image pre-training with frozen image encoders and large language models},
  author={Li, Junnan and Li, Dongxu and Savarese, Silvio and Hoi, Steven},
  booktitle={International conference on machine learning},
  pages={19730--19742},
  year={2023},
  organization={PMLR}
}

@inproceedings{hirota2022quantifying,
  title={Quantifying Societal Bias Amplification in Image Captioning},
  author={Hirota, Yusuke and Nakashima, Yuta and Garcia, Noa},
  booktitle={Proceedings of the IEEE/CVF Conference on Computer Vision and Pattern Recognition},
  pages={13440--13449},
  year={2022}
}

@inproceedings{berg2022prompt,
  title={A Prompt Array Keeps the Bias Away: Debiasing Vision-Language Models with Adversarial Learning},
  author={Berg, Hugo and Hall, Siobhan and Bhalgat, Yash and Kirk, Hannah andShtedritski, Aleksandar and Bain, Max},
  booktitle={Proceedings of the 2nd Conference of the Asia-Pacific Chapter of the Association for Computational Linguistics and the 12th International Joint Conference on Natural Language Processing (Volume 1: Long Papers)},
  pages={806--822},
  year={2022}
}

@inproceedings{Seth2023DeAR,
  title={DeAR: Debiasing Vision-Language Models with Additive Residuals},
  author={Ashish Seth and Mayur Hemani and Chirag Agarwal},
  year={2023},
  pages={6820-6829},
  booktitle={Proceedings of the IEEE/CVF Conference on Computer Vision and Pattern Recognition}
}

@inproceedings{zhao2017men,
  title={Men Also Like Shopping: Reducing Gender Bias Amplification using Corpus-level Constraints},
  author={Zhao, Jieyu and Wang, Tianlu and Yatskar, Mark and Ordonez, Vicente and Chang, Kai-Wei},
  booktitle={Proceedings of the 2017 Conference on Empirical Methods in Natural Language Processing},
  pages={2979--2989},
  year={2017}
}

@InProceedings{Hendricks2018Women,
  author={Hendricks, Lisa Anne and Burns, Kaylee and Saenko, Kate and Darrell, Trevor and Rohrbach, Anna},
  title={Women also Snowboard: Overcoming Bias in Captioning Models},
  booktitle={Proceedings of the European Conference on Computer Vision},
  year={2018}
}

@article{friedrich2023fairdiffusion,
  title={Fair Diffusion: Instructing Text-to-Image Generation Models on Fairness}, 
  author={Felix Friedrich and Manuel Brack and Lukas Struppek and Dominik Hintersdorf and Patrick Schramowski and Sasha Luccioni and Kristian Kersting},
  year={2023},
  journal={arXiv preprint arXiv:2302.10893}
}

@article{sahili2025faircot,
  title={FairCoT: Enhancing Fairness in Text-to-Image Generation via Chain of Thought Reasoning with Multimodal Large Language Models}, 
  author={Zahraa Al Sahili and Ioannis Patras and Matthew Purver},
  year={2025},
  journal={arXiv preprint arXiv:2406.09070}
}

@inproceedings{kim2023destereotyping,
  title={De-stereotyping Text-to-Image Models through Prompt Tuning},
  author={Eunji Kim and Siwon Kim and Chaehun Shin and Sungroh Yoon},
  booktitle={ICML 2023 Workshop on Challenges in Deployable Generative AI},
  year={2023}
}

@InProceedings{Gandikota2024concept,
  author= {Gandikota, Rohit and Orgad, Hadas and Belinkov, Yonatan and Materzy\'nska, Joanna and Bau, David},
  title= {Unified Concept Editing in Diffusion Models},
  booktitle={Proceedings of the IEEE/CVF Winter Conference on Applications of Computer Vision},
  year={2024},
  pages={5111-5120}
}

@InProceedings{li2024selfdiscovery,
  author={Li,Hang and Shen,Chengzhi and Torr,Philip and Tresp,Volker and Gu,Jindong},
  title={Self-Discovering Interpretable Diffusion Latent Directions for Responsible Text-to-Image Generation},
  booktitle={Proceedings of the IEEE/CVF Conference on Computer Vision and Pattern Recognition},
  year={2024},
  pages={12006--12016}
}

@article{li2025fairmapping,
    title={Fair Text-to-Image Diffusion via Fair Mapping},
    volume={39},
    number={25},
    journal={Proceedings of the AAAI Conference on Artificial Intelligence},
    author={Li,Jia and Hu,Lijie and Zhang,Jingfeng and Zheng,Tianhang and Zhang,Hua and Wang,Di},
    year={2025},
    pages={26256--26264}
}

@article{shen2024finetuning,
  title={Finetuning Text-to-Image Diffusion Models for Fairness}, 
  author={Xudong Shen and Chao Du and Tianyu Pang and Min Lin and Yongkang Wong and Mohan Kankanhalli},
  year={2024},
  journal={arXiv preprint arXiv:2311.07604}
}

@InProceedings{Zhang2023iti,
  author={Zhang,Cheng and Chen,Xuanbai and Chai,Siqi and Wu,Chen Henry and Lagun,Dmitry and Beeler,Thabo and De la Torre,Fernando},
  title={ITI-GEN: Inclusive Text-to-Image Generation},
  booktitle={Proceedings of the IEEE/CVF International Conference on Computer Vision},
  year={2023},
  pages={3969--3980}
}

@article{narayanan2025bias,
  title={Bias in the Picture: Benchmarking VLMs with Social-Cue News Images and LLM-as-Judge Assessment}, 
  author={Aravind Narayanan and Vahid Reza Khazaie and Shaina Raza},
  year={2025},
  journal={arXiv preprint arXiv:2509.19659}
}

@article{ganguli2023capacity,
  title={The capacity for moral self-correction in large language models},
  author={Ganguli, Deep and Askell, Amanda and Schiefer, Nicholas and Liao, Thomas I and Luko{\v{s}}i{\=u}t{\.e}, Kamil{\.e} and Chen, Anna and Goldie, Anna and Mirhoseini, Azalia and Olsson, Catherine and Hernandez, Danny and others},
  journal={arXiv preprint arXiv:2302.07459},
  year={2023}
}

@article{pan2024automatically,
  title={Automatically correcting large language models: Surveying the landscape of diverse automated correction strategies},
  author={Pan, Liangming and Saxon, Michael and Xu, Wenda and Nathani, Deepak and Wang, Xinyi and Wang, William Yang},
  journal={Transactions of the Association for Computational Linguistics},
  volume={12},
  pages={484--506},
  year={2024},
  publisher={MIT Press One Broadway, 12th Floor, Cambridge, Massachusetts 02142, USA~…}
}

@inproceedings{liu2024intrinsic,
  title={Intrinsic self-correction for enhanced morality: An analysis of internal mechanisms and the superficial hypothesis},
  author={Liu, Guangliang and Mao, Haitao and Tang, Jiliang and Johnson, Kristen},
  booktitle={Proceedings of the 2024 Conference on Empirical Methods in Natural Language Processing},
  pages={16439--16455},
  year={2024}
}

@article{reinhard2002color,
  title={Color transfer between images},
  author={Reinhard, Erik and Adhikhmin, Michael and Gooch, Bruce and Shirley, Peter},
  journal={IEEE Computer graphics and applications},
  volume={21},
  number={5},
  pages={34--41},
  year={2002},
  publisher={IEEE}
}

@article{bolukbasi2016man,
  title={Man is to computer programmer as woman is to homemaker? debiasing word embeddings},
  author={Bolukbasi, Tolga and Chang, Kai-Wei and Zou, James Y and Saligrama, Venkatesh and Kalai, Adam T},
  journal={Advances in neural information processing systems},
  volume={29},
  year={2016}
}

@inproceedings{ke2021musiq,
  title={Musiq: Multi-scale image quality transformer},
  author={Ke, Junjie and Wang, Qifei and Wang, Yilin and Milanfar, Peyman and Yang, Feng},
  booktitle={Proceedings of the IEEE/CVF international conference on computer vision},
  pages={5148--5157},
  year={2021}
}

@inproceedings{lin2024towards,
  title={Towards counterfactual fairness-aware domain generalization in changing environments},
  author={Lin, Yujie and Zhao, Chen and Shao, Minglai and Meng, Baoluo and Zhao, Xujiang and Chen, Haifeng},
  booktitle={Proceedings of the Thirty-Third International Joint Conference on Artificial Intelligence},
  pages={4560--4568},
  year={2024}
}

@inproceedings{lin2025fade,
  title={FADE: towards fairness-aware data generation for domain generalization via classifier-guided score-based diffusion models},
  author={Lin, Yujie and Li, Dong and Shao, Minglai and Wan, Guihong and Zhao, Chen},
  booktitle={Proceedings of the Thirty-Fourth International Joint Conference on Artificial Intelligence},
  pages={439--447},
  year={2025}
}

@inproceedings{
lin2026bidirectional,
title={Bi-directional Bias Attribution: Debiasing Large Language Models without Modifying Prompts},
author={Yujie Lin and Kunquan Li and YiXuan Liao and Xiaoxin Chen and Jinsong Su},
booktitle={The Fourteenth International Conference on Learning Representations},
year={2026},
url={https://openreview.net/forum?id=mUTN9VIaSy}
}

@inproceedings{
lin2026zerounlearn,
title={ZeroUnlearn: Few-Shot Knowledge Unlearning in Large Language Models},
author={Yujie Lin and Chengyi Yang and Zhishang Xiang and Yiping Song and Jinsong Su},
booktitle={Forty-third International Conference on Machine Learning},
year={2026},
url={https://openreview.net/forum?id=vvxcADnfL9}
}

@article{jia2026object,
  title={Object Hallucination-Free Reinforcement Unlearning for Vision-Language Models},
  author={Jia, Kaidi and Lin, Yujie and Yang, Chengyi and Ma, Jiayao and Su, Jinsong},
  journal={arXiv preprint arXiv:2605.08031},
  year={2026}
}

\end{document}